\documentclass{article}

\usepackage{microtype}
\usepackage{graphicx}
\usepackage{subfigure}
\usepackage{booktabs}

\usepackage{hyperref}

\usepackage[accepted]{icml2025}

\usepackage{amsmath}
\usepackage{amssymb}
\usepackage{mathtools}
\usepackage{amsthm}

\usepackage{wrapfig}

\usepackage{listings}
\usepackage[capitalize,noabbrev]{cleveref}

\usepackage[utf8]{inputenc} 
\usepackage[T1]{fontenc}    
\usepackage{hyperref}       
\usepackage{url}            
\usepackage{booktabs}       
\usepackage{amsfonts}       
\usepackage{nicefrac}       
\usepackage{microtype}      
\usepackage{xcolor}         
\usepackage{pgfplots}
\usepackage{subcaption}
\usepackage{wrapfig}
\pgfplotsset{compat=1.18}
\usepgfplotslibrary{groupplots}

\theoremstyle{plain}

\theoremstyle{definition}

\theoremstyle{remark}

\usepackage[textsize=tiny]{todonotes}

\icmltitlerunning{Random Initialization of Gated Sparse Adapters}

\begin{document}

\twocolumn[
\icmltitle{Random Initialization of Gated Sparse Adapters}

\begin{icmlauthorlist}
\icmlauthor{Vi Retault}{poly}
\icmlauthor{Yohaï-Eliel Berreby}{phgy,mila}
\end{icmlauthorlist}

\icmlaffiliation{poly}{GIGL Department, Polytechnique Montréal, Montréal, QC, Canada}
\icmlaffiliation{phgy}{Department of Physiology, McGill University, Montréal, QC, Canada}
\icmlaffiliation{mila}{Mila - Quebec AI Institute}

\icmlcorrespondingauthor{Vi Retault}{vi@retau.lt}
\icmlcorrespondingauthor{Yohaï-Eliel Berreby}{me@yberreby.com}

\icmlkeywords{Large Language Models, Sparsity}

\vskip 0.3in
]

\printAffiliationsAndNotice{}  

\begin{abstract}
When fine-tuning language models on new tasks, catastrophic forgetting --- performance degradation on previously-learned tasks --- is a ubiquitous problem.
While Parameter-Efficient Fine-Tuning (PEFT) methods like LoRA address this through low-rank adapters, sparse adaptation offers an alternative that doesn't impose rank constraints.
We introduce Random Initialization of Gated Sparse Adapters (RIGSA), which starts from randomly-initialized full-rank adapters, gates them with a ReZero analog, and sparsifies them with iterative magnitude pruning.
We evaluate RIGSA on SmolLM2-1.7B-Instruct using a novel vision-in-text task (Textual MNIST) and measure forgetting on PIQA, HellaSwag, and GSM8k.
SmolLM2-1.7B-Instruct initially performs around chance level on Textual MNIST, and is capable of learning the task through RIGSA, 4-bit QLoRA and random masking.
In spite of having more trainable parameters than QLoRA, the RIGSA configurations that we studied displayed less forgetting than QLoRA, particularly on GSM8k, though it performs comparably to random masking.
\end{abstract}

\section{Introduction}
\label{introduction}

In recent years, the field of deep learning has undergone a fundamental shift in how models are developed and trained.
Rather than designing task-specific architectures trained from scratch on labeled datasets, the prevailing approach now centers around \textit{foundation models}~\citep{bommasani2021opportunities} --- large neural networks pre-trained on extensive, diverse data using self-supervised objectives.
These models serve as general-purpose representations of the training data that can be adapted to a wide range of downstream tasks~\citep{raffel2020exploring}, and can achieve strong performance even on tasks where labeled data is limited~\citep{devlin2018bert}.

Perhaps the most prominent category of foundation models is Large Language Models (LLMs), which are trained on massive text corpora collected from the Internet.
These models can be adapted to different tasks through prompting~\citep{radford2019language} --- a technique where the model receives a natural-language input specifically tailored to elicit the desired behavior --- and have demonstrated remarkable capabilities in question answering~\citep{touvron2023llama}, zero-shot reasoning~\citep{kojima2022large}, and learning from in-context examples~\citep{brown2020language}.

To fully leverage the potential of foundation models, researchers often fine-tune them on task-specific datasets to improve performance beyond what prompting alone can achieve~\citep{ouyang2022training, chung2024scaling}.
Fine-tuning approaches generally fall into two categories: full fine-tuning, which modifies all model weights, and Parameter-Efficient Fine-Tuning (PEFT), where only a subset of the parameters is updated~\citep{fu2023effectiveness}.
The latter category has become essential for working with large-scale models~\citep{ding2023parameter}, as it significantly reduces computational and storage requirements~\citep{hu2021lora, heetal2022sparseadapter}, can mitigate catastrophic forgetting~\citep{houlsby2019parameter, wang2023orthogonal}, and enables composition of multiple adapters~\citep{pfeifferetal2021adapterfusion}.

Low-Rank Adaptation (LoRA)~\citep{hu2021lora} is a PEFT technique that has gained significant traction in the research community~\citep{mao2025survey}, and is widely used in practice.
Given an initial, frozen weight matrix $W_0 \in \mathbb{R}^{d \times k}$, LoRA learns a low-rank difference matrix $\Delta W = AB$ where $A \in \mathbb{R}^{d \times r}$ and $B \in \mathbb{R}^{r \times k}$. The rank $r \ll \min(d, k)$ of $\Delta W$ controls the expressiveness of the adapter when applied to the original weight matrix, such that $W = W_0 + \Delta W$. Intuitively, $\Delta W$ modifies how $W_0$ responds to components of the input vector that lie in the low-dimensional hyperplane $\ker(B)^{\perp}$ by adding an output vector in $\mathrm{im}(A)$.

Despite some initial results suggesting LoRA can match full fine-tuning performance for some classic NLP tasks~\citep{hu2021lora, zhao2024lora}, the low-rank constraint imposes important limitations. Empirical studies show LoRA underperforms full fine-tuning on more complex tasks like mathematical reasoning and programming~\citep{biderman2024lora}, and exhibits reduced out-of-distribution robustness when the rank is low~\citep{shuttleworth2024lora}.

Sparse fine-tuning offers an alternative approach to developing memory-efficient adapters by training a sparse matrix $\Delta W_s$ such that $W = W_0 + \Delta W_s$~\citep{song2024sparse}. Unlike LoRA, sparse fine-tuning techniques do not impose a hard constraint on the rank of the resulting update, potentially allowing for more expressive adaptations. A key challenge in this approach lies in identifying the sparsity mask $m$ for $\Delta W_s$, which determines which weights of the pre-trained model $W_0$ will be modified during training. Several methods exist for identifying this mask, including approaches based on accumulated gradients~\citep{ansell2024scaling}, iterative pruning during training~\citep{guo2020parameter}, second-order approximations~\citep{fu2023effectiveness}, and the Fisher information matrix~\citep{xuetal2021raise}.

A promising, yet relatively unexplored approach to identifying the sparsity mask $m$ is to leverage the Lottery Ticket Hypothesis (LTH)~\citep{frankle2018lottery}, which allows finding sparse sub-networks within dense neural networks that can match their accuracy. Building on this principle, Lottery Ticket Sparse Fine-Tuning (LT-SFT)~\citep{anselletal2022composable} employs a two-stage process: it first performs full fine-tuning of a pre-trained language model on the target task and identifies the weights that changed the most, then resets the model to its pre-trained state and trains only those weights. The authors successfully applied this method to construct sparse, composable adapters for cross-lingual transfer.

\textbf{Contributions}: In this work, we propose Random Initialization of Gated Sparse Adapters (RIGSA), a generalization of LT-SFT supporting non-zero initializations of the difference matrix $\Delta W$, as well as multiple pruning steps.
We evaluate our approach on a state-of-the-art 1.7B-parameter language model, SmolLM2-1.7B-Instruct~\citep{allal2025smollm2}.
We introduce a challenging out-of-distribution vision-in-text task, Textual MNIST, to rigorously evaluate our method.

\section{Related Work}
\label{related_work}

\subsection{Lottery Ticket Hypothesis}
The Lottery Ticket Hypothesis (LTH)~\citep{frankle2018lottery} states that a randomly initialized, over-parameterized dense neural network contains sparse sub-networks called \textit{winning tickets} that can be trained in isolation with a similar number of iterations to achieve the same accuracy as the dense network. In practice, a dense neural network $W_0$ is randomly initialized with a certain distribution $\mathcal{D}$,
and trained on a dataset for $j$ iterations. Its weights of lower magnitude after training are pruned with a binary mask $m \in \{0,1\}^{|W|}$, and its weights of largest magnitude are reset to their initialization value. The resulting sparse network $m \odot W_0$ is trained on the same dataset for $j' \approx j$ iterations, and should perform comparably to the dense network. This process can be repeated to create increasingly sparse networks, a technique known as Iterative Magnitude Pruning (IMP)~\citep{frankle2018lottery}.

The original authors initially provided empirical evidence of the Lottery Ticket Hypothesis on small fully-connected and convolutional neural networks, but follow-up work managed to find winning tickets for larger networks and more complex architectures, including Transformers~\citep{vaswani2017attention}, by resetting the weights to their values after several training iterations rather than at initialization~\citep{frankle2020linear, brixetal2020successfully}. Additionally,~\citet{zhou2019deconstructing} developed more effective sparsity masks by targeting weights moving toward zero, while~\citet{You2020Drawing} introduced efficiency improvements through early termination of the full fine-tuning step.

\subsection{Sparse Adaptation}
Several recent studies have proposed sparse adaptation techniques for LLMs.

Lottery Ticket Sparse Fine-Tuning (LT-SFT)~\citep{anselletal2022composable} adapts BERT~\citep{devlin2018bert} to new languages by training a dense difference matrix $\Delta W$, pruning the parameters of lowest magnitude, and training again from initialization. While this method is inspired by the Lottery Ticket Hypothesis, it initializes $\Delta W$ to zero rather than randomly, which cannot result in a "lucky" initialization of one of its sparse sub-networks.

To avoid the costly full fine-tuning phase, Sparse Increment Fine-Tuning (SIFT)~\citep{song2024sparse} updates only the parameters with the highest estimated gradient magnitudes, computed on the first batch of training data. The authors demonstrate that their method achieves performance competitive with full fine-tuning and Low-Rank Adaptation (LoRA) on language understanding and Python programming tasks, while updating fewer parameters. Interestingly, they note that choosing random masks also leads to non-trivial performance.

SpIEL~\citep{ansell2024scaling} alternates growing and shrinking the sparse adaptation matrix $\Delta W_s$ during training, dropping low-magnitude weights and adding weights with the highest estimated gradient magnitudes. Despite resulting in similar speed and memory efficiency as LoRA, this method requires using a dynamic sparsity mask, updated throughout training.

Robust Adaptation (RoSA)~\citep{nikdan2024rosa} combines sparse adaptation with LoRA, outperforming both techniques through joint training of two adapters $\Delta W = AB$ and $\Delta W_s$. Similarly to~\citet{ansell2024scaling}, the RoSA authors select the sparsity mask of $\Delta W_s$ using the magnitude of accumulated gradients computed on a fraction of the training dataset. They observe that finding the mask using information obtained after full fine-tuning, similarly to the Lottery Ticket Hypothesis, performs even better, and that using a random mask already outperform LoRA on mathematical reasoning.

\section{Our Approach}

\subsection{Initial Problem Statement}

Given a foundation model with base parameters $W_0$ optimized for a source task $\mathcal{T}_S$, the \textit{adaptation problem} involves finding a difference matrix $\Delta W$ such that $W = W_0 + \Delta W$ obtains a high accuracy on a target task $\mathcal{T}_T$, while retaining high accuracy on the source task $\mathcal{T}_S$.
In this setting, we consider $W_0$ to be a fixed part of the architecture, and only optimize the difference parameters $\Delta W$. 

Drawing parallels with the Lottery Ticket Hypothesis perspective, one could view $\Delta W$ as a network.
Within it, is it possible to find a sparse sub-network $\Delta W_s$ capable of being trained in isolation to achieve comparable accuracy to $\Delta W$ on both the target and source tasks?

In this paper, we focus on cases in which $\Delta W$ is allowed to be full-rank, and is not simply initialized to $0$.
This sets our approach apart from RoSA~\citep{nikdan2024rosa}, which considers low-rank difference matrices, and LT-SFT~\citep{anselletal2022composable}.

\subsection{IMP on Randomly-Initialized, Gated Difference Matrices}

To find a ticket, we could consider training $W = W_0 + \Delta W$ on the target task, with $\Delta W$ dense and randomly initialized to allow the possibility of containing a winning ticket, and subsequently pruning $\Delta W$. However, starting the optimization process from such a random perturbation of $W_0$ may introduce instabilities during early training and hurt performance on the source tasks. LoRA~\citep{hu2021lora} addresses this issue by initializing $B$ (but not $A$) to 0, to start precisely from $W_0$.
We draw inspiration from this strategy in our approach, without parameterizing $\Delta W$ as a product of low-rank matrices.
We borrow a trick from ReZero~\citep{bachlechner2021rezero}: multiplication with a single (near-)zero-initialized learnable parameter, termed $\alpha$.
By optimizing $W = W_0 + \alpha \Delta W$, we start the optimization process from the pre-trained weights $W_0$ --- because $\alpha$ is initialized at or near zero --- but allow it to deviate from those weights --- because $\Delta W$ is randomly initialized --- and then come back in that region due to weight decay. In practice, we initialize $\alpha$ to a small positive value ($10^{-6}$).

After training, we follow the advances of~\citet{zhou2019deconstructing} and apply pruning to applicable parameters of $\Delta W$ (all linear layers except the language modelling head). 
We reset to their initialization value a certain ratio of the weights that didn't change sign (sorted to keep those with the largest final magnitude), setting the others to zero and freezing them.
We call the resulting sparse matrix $\Delta W_1$, and train it again on the target task. We repeat this process iteratively, following the Iterative Magnitude Pruning (IMP)~\citep{frankle2018lottery} technique, to find increasingly sparser $\Delta W_i$. After a certain fixed number of steps, we obtain a randomly initialized sparse $\Delta W_T$, our winning ticket, that we train one last time.

\section{Methodology}

\subsection{Base model}

We choose SmolLM2-1.7B-Instruct~\citep{allal2025smollm2} as our base model, with all of its parameters $W_0$ frozen. SmolLM2 is a state-of-the-art language model with 1.7 billion parameters developed by Hugging Face and trained on approximately 11 trillion tokens.
This model was selected because it is small enough to run fine-tuning experiments on our limited hardware, while exhibiting non trivial performance on complex tasks. Table~\ref{tab:smollm2_baseline} shows its baseline performance as reported by the authors, which serves as a reference point for experiments.
Note that we are calling this our \emph{base} model, even though it was instruct-tuned by its authors, because we perform further fine-tuning to learn the target task.

\begin{table}
  \caption{Performance of SmolLM2-1.7B-Instruct on benchmarks, as reported in~\citet{allal2025smollm2}.}
  \label{tab:smollm2_baseline}
  \centering
  \begin{tabular}{ll}
    \toprule
    \textbf{Benchmark} & \textbf{Performance} \\
    \midrule
    PIQA & 74.4 \\
    HellaSwag & 66.1 \\
    GSM8k (5-shot) & 48.8 \\
    \bottomrule
  \end{tabular}
\end{table}

\subsection{Target Task: Textual MNIST}
\label{vision_in_text}

What target task should we learn with our adapter? Ideally, it should be both out-of-distribution, so it differs from the source tasks, and standardized, so results are comparable across models. Common LLM benchmarks typically involve NLP tasks such as text classification, sentiment analysis, or question answering. However, SmolLM2 already performs well on such tasks, for instance achieving 66.1\% zero-shot accuracy on the sentence completion benchmark HellaSwag~\citep{allal2025smollm2}. To meaningfully challenge the model, we introduce a vision-in-text image classification task based on MNIST, and confirm that SmolLM2 initially struggles with it.

We consider MNIST~\citep{lecun1998gradient} as our target task because it is widely used in previous LTH studies~\citep{frankle2018lottery, zhou2019deconstructing}, and represents a clear distribution shift from the data used to pre-train language models. This choice helps reduce the influence of transfer learning and provides a clear measure of the model's ability to learn a novel target task.

\begin{figure*}
  \centering
  \begin{minipage}{0.4\textwidth}
    \includegraphics[width=\textwidth]{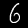}
  \end{minipage}
    \hspace{50.0pt}
  \begin{minipage}{0.22\textwidth}
    \ttfamily\tiny
    \begin{verbatim}
0000000000000000000000000000
0000000000000000000000000000
0000000000000000000000000000
0000000000000000169100000000
0000000000000002985000000000
0000000000000029500000000000
0000000000000197000000000000
0000000000000770000000000000
0000000000005910000000000000
0000000000009700000000000000
0000000000049100000000000000
0000000000077000000000000000
0000000000291000210000000000
0000000000580039994000000000
0000000000930083039200000000
0000000001930250005700000000
0000000002910000001900000000
0000000002900000000900000000
0000000002900000003900000000
0000000002900000007500000000
0000000000840000068000000000
0000000000494013881000000000
0000000000049999500000000000
0000000000000000000000000000
0000000000000000000000000000
0000000000000000000000000000
0000000000000000000000000000
0000000000000000000000000000
    \end{verbatim}
  \end{minipage}
  \caption{
  An MNIST image of the digit 6 (left), and its Textual MNIST representation (right).
  0-255 pixel values are quantized to the 0-9 range, so that each pixel maps to a single ASCII digit.
  }
  \label{fig:textual_mnist}
\end{figure*}

MNIST is a popular image classification dataset composed of 60,000 training and 10,000 test greyscale images (28$\times$28 pixels) of handwritten digits (0-9).
Because language models operate on text, we convert each image into a textual format. We draw inspiration from BigBench~\citep{srivastava2022beyond}, which uses ASCII art to represent MNIST digits for evaluating LLMs on image recognition in text. Rather than using ASCII art, we quantize the greyscale value of each pixel to the range $[0, 9]$ and represent the image as rows of digit sequences, separated by newlines, as illustrated in Figure~\ref{fig:textual_mnist}. This format aligns well with the tokenizer of SmolLM2, which splits digits individually, ensuring a consistent sequence length across images.
In all of our experiments, we reserve 1\% of the training set for validation, and we follow the original MNIST train-test split to construct the test set of Textual MNIST.

\begin{table}
  \centering
  \caption{Test accuracy of SmolLM2-1.7B-Instruct on Textual MNIST.}
  \label{tab:smollm2_mnist_baseline}
  \begin{tabular}{lcc}
    \toprule
    \textbf{Variant}   & \textbf{0-shot} & \textbf{5-shot} \\
    \midrule
    Full Precision     & 9.16               & 10.04          \\
    Quantized (4-bit)    & 9.80               & 10.41             \\
    \bottomrule
  \end{tabular}
\end{table}

Recent work has shown that compact language models perform poorly on vision-in-text-benchmarks~\citep{jia2024visual,wang2024bot,jiang2024artprompt}, supporting our choice of this task as out-of-distribution for the pre-trained model. To confirm this, we evaluate the zero-shot and 5-shot test accuracy of SmolLM2-1.7B-Instruct on Textual MNIST. The Instruct variant has already been fine-tuned by its authors to follow natural language instructions, allowing us to prompt the model directly using the ChatML format~\citep{openaichatml}, as shown in Figure~\ref{fig:textual_mnist_prompt}. To standardize output parsing, we append the prefix "The digit is " to the assistant's response, and constrain generation to a single token. In the 5-shot setting, we also include 5 examples with their labels from the training set in context, inserted as previous conversation turns. 

We run this evaluation using both the full-precision model and a 4-bit quantized version from Unsloth~\citep{unsloth}, commonly used to speed up inference, and present the results in Table~\ref{tab:smollm2_mnist_baseline}. In both cases, the near-random test accuracy supports that the task is out-of-distribution for the model. The slight accuracy drop below random guessing (10\%) in the zero-shot setting reflects the occasional failure to produce a valid digit as the next token. In the 5-shot setting, the model always generates a digit, but still doesn't perform much better than chance.

\subsection{Source Tasks: PIQA, HellaSwag, GSM8k}
\label{source_tasks}

We select 3 source tasks on which SmolLM2 is already proficient, and quantify how performance degrades on those tasks after fine-tuning. Table~\ref{tab:smollm2_baseline} reports the performance of three common LLM benchmarks on which SmolLM2 is competitive with other models of similar size~\citep{allal2025smollm2}, Physical Interaction: Question Answering (PIQA)~\citep{Bisk2020}, Harder Endings, Longer contexts, and Low-shot Activities for Situations With Adversarial Generations (HellaSwag)~\citep{zellers2019hellaswag}, and Grade School Math 8K (GSM8k)~\citep{cobbe2021gsm8k}. PIQA is a physical commonsense reasoning dataset with a test set of 3,000 questions describing everyday situations, each matched with two possible answers. HellaSwag is a sentence completion dataset with a test set containing 10K beginning of sentences, with four possible endings each. GSM8k is a mathematical reasoning dataset with a validation set of 1319 math problems with numerical answers.

\subsection{Pruning Procedure}

We perform all pruning experiments on a single RTX 4090, which has 24 GB of VRAM.
We use the AdamW~\citep{loshchilov2018decoupled} optimizer with a very high weight decay ($1.0$) to encourage sparsity.
We follow a piecewise-linear learning rate schedule: 1000 warmup steps peaking at $2 \times 10^{-3}$ followed by decay to zero. 
Since only a single training sample can fit in GPU memory, we use gradient accumulation to obtain an effective batch size of $16$.
At each IMP iteration, we train our adapter for one epoch.
Then, we follow the criteria laid out above: among free parameters that did not change sign relative to their initial values, we keep the 80 \% with the largest magnitude, and reset the remaining parameters to their initial values.
We stop after 5 iterations, resulting in a sparse adapter with $3.46 \%$ of the original trainable parameters.
As a baseline, we also train an adapter with a random mask of that same sparsity ratio.

\section{Experiments}
\label{winning_ticket}

\begin{table}
  \caption{Accuracy of SmolLM2-1.7B-Instruct on benchmarks.}
  \label{tab:smollm2_forgetting_baseline}
  \centering
  \begin{tabular}{lcc}
    \toprule
      \textbf{Benchmark} & \textbf{Full Precision} & \textbf{Quantized (4-bit)} \\
    \midrule
    PIQA                 & 75.4 & 75.2  \\
    HellaSwag            & 51.7 & 50.8  \\
    GSM8k (5-shot)       & 43.7 & 19.9  \\
    \bottomrule
  \end{tabular}
\end{table}

\begin{figure*}[!h]
  \begin{minipage}{0.48\linewidth}
    \begin{tikzpicture}
      \begin{axis}[
        title={\textbf{a. Textual MNIST (target)}},
        xlabel={Pruning Iteration}, xtick={1,2,3,4,5},
        ylabel={Accuracy (\%)},
        ymin=98, ymax=100,
        ytick={98.0,99.0,100},
        yticklabel style={
          /pgf/number format/fixed,       
          minimum width=3em,              
          align=right                     
        },
        reverse legend,
        width=\textwidth,
        legend cell align={left},
      ]
      
        \addplot[black,line width=2pt,dash pattern=on 0.5pt off 8pt, line cap=round]             coordinates {(1,98.31) (5,98.31)};
         \addplot[blue,very thick,mark=*]    coordinates {(1,99.05) (2,98.85) (3,98.58) (4, 98.33) (5, 98.37)};
        \legend{Random Mask, RIGSA}
      \end{axis}
    \end{tikzpicture}
  \end{minipage}
  \begin{minipage}{0.48\linewidth}
    \begin{tikzpicture}
      \begin{axis}[
        title={\textbf{b. GSM8k (source)}},
        xlabel={Pruning Iteration}, xtick={1,2,3,4,5},
        ylabel={Accuracy (\%)},
        ymin=40, ymax=50,
        legend pos=north east,
        reverse legend,
        legend style={font=\footnotesize},
        width=\textwidth,
      ]
       
        \addplot[black,line width=2pt,dash pattern=on 0.5pt off 8pt, line cap=round]             coordinates {(1,44.58) (5,44.58)};
        \addplot[black,thick,dash pattern=on 10pt off 5pt]             coordinates {(1,43.75) (5,43.75)};
        \addplot[red,very thick,mark=*] coordinates {(1,40.71) (2,44.42) (3,45.11) (4,44.50) (5,44.05)};
        \legend{Random Mask, Baseline, RIGSA}
      \end{axis}
    \end{tikzpicture}
  \end{minipage}\vspace{1em}
  \begin{minipage}{0.48\linewidth}
    \begin{tikzpicture}
      \begin{axis}[
        title={\textbf{c. HellaSwag (source)}},
        xlabel={Pruning Iteration}, xtick={1,2,3,4,5},
        ylabel={Accuracy (\%)},
        ymin=50, ymax=54,
        yticklabel style={
          /pgf/number format/fixed,       
          minimum width=3em,              
          align=right                     
        },
        legend pos=north east,
        reverse legend,
        legend style={font=\footnotesize},
        width=\textwidth,
      ]
        \addplot[black,line width=2pt,dash pattern=on 0.5pt off 8pt, line cap=round]             coordinates {(1,52.20) (5,52.20)};
        \addplot[black,thick,dash pattern=on 10pt off 5pt]                coordinates {(1,51.71) (5,51.71)};
        \addplot[orange,very thick,mark=*] coordinates {(1,51.55) (2,51.80) (3,51.57) (4,51.69) (5,51.74)};
        \legend{Random Mask, Baseline, RIGSA}
      \end{axis}
    \end{tikzpicture}
  \end{minipage}\hfill
  \begin{minipage}{0.48\linewidth}
    \begin{tikzpicture}
      \begin{axis}[
        title={\textbf{d. PIQA (source)}},
        xlabel={Pruning Iteration}, xtick={1,2,3,4,5},
        ylabel={Accuracy (\%)},
        ymin=74, ymax=77,
        legend pos=north east,
        reverse legend,
        legend style={font=\footnotesize},
        width=\textwidth,
      ]
        \addplot[black,line width=2pt,dash pattern=on 0.5pt off 8pt, line cap=round]             coordinates {(1,75.57) (5,75.57)};
        \addplot[black,thick,dash pattern=on 10pt off 5pt]               coordinates {(1,75.41) (5, 75.41)};
        \addplot[green!60!black,very thick,mark=*] coordinates {(1,75.68) (2,75.19) (3,75.46) (4,75.24) (5,75.19)};
        \legend{Random Mask,Baseline,RIGSA}
      \end{axis}
    \end{tikzpicture}
  \end{minipage}
  \caption{
  Test accuracy on source and target tasks at each pruning iteration, for our method ("RIGSA"), a random mask with a parameter budget equal to the one available at the last pruning iteration ("Random Mask"), and the model before any fine-tuning ("Baseline").
  \textbf{a.}: On the target task (Textual MNIST), more pruning iterations (and thus a smaller number of trainable parameters) leads to an accuracy drop, ultimately reverting to near the random-mask performance.
  \textbf{b-d.}:
Fine-tuning with a random mask appears to improve performance on some source tasks, though this may be due to inter-run variability rather than a genuine effect, as one would typically expect performance to degrade (or remain unchanged) on source tasks when fine-tuning on a new target task.
  }
  \label{fig:smollm2_forgetting_step}
\end{figure*}
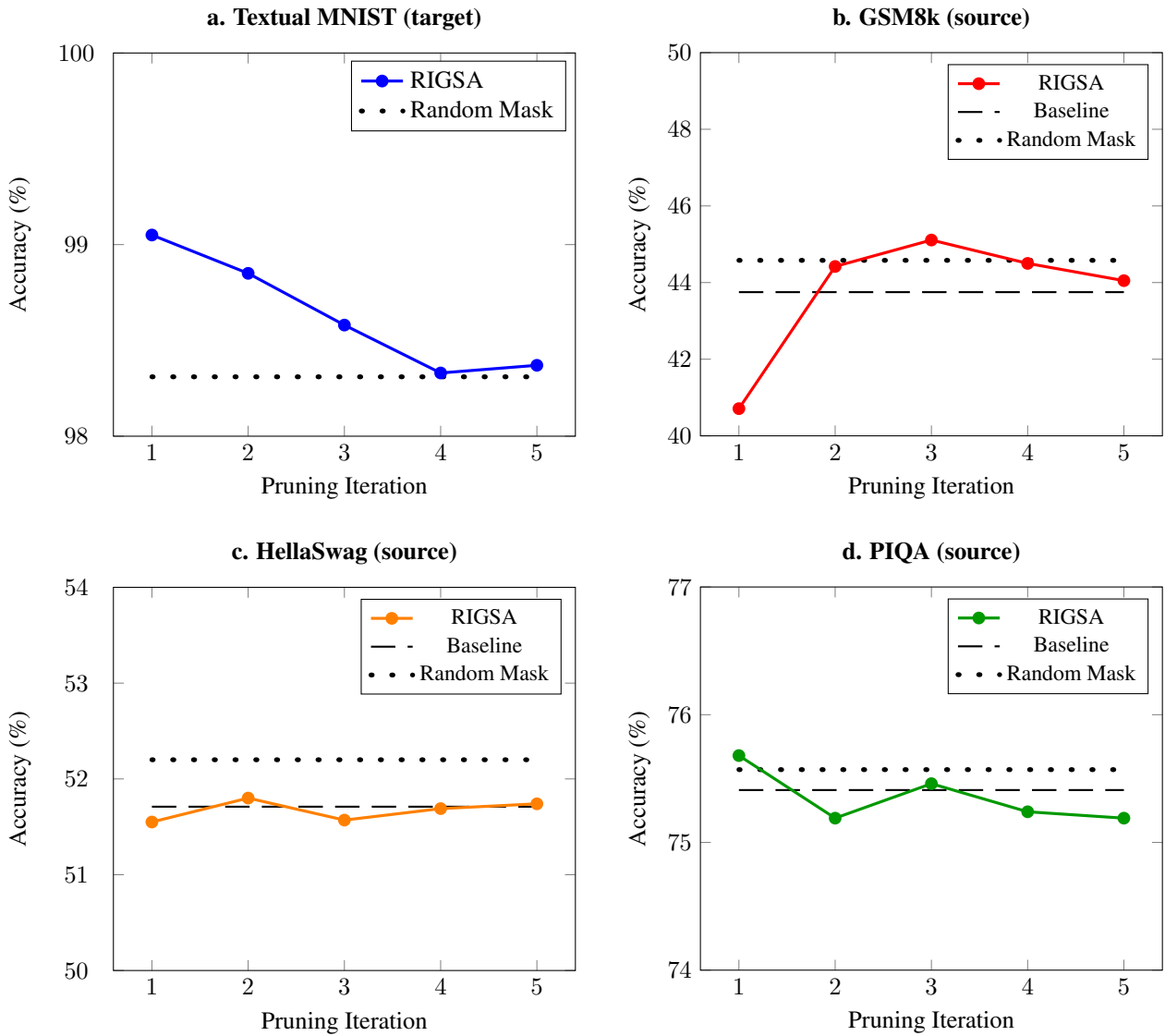

\begin{figure*}[!h]
  \begin{minipage}{0.48\linewidth}
   \begin{tikzpicture}
  \begin{axis}[
    title={\textbf{a. Textual MNIST (target)}},
    xlabel={Rank},
    xtick={1,4,8,16},
    ylabel={Accuracy (\%)},
    ymode=log,
    ymin=1,
    ymax=600,
    ytick={1,10,100},
    yticklabels={1,10,100},
    legend pos=north east,
    legend style={font=\footnotesize},
    width=\textwidth,
    yticklabel style={
      /pgf/number format/fixed,       
      minimum width=3em,              
      align=right                     
    },
  ]
    \addplot[blue,very thick,mark=*]
      coordinates {(1,99.45) (4,99.43) (8,99.41) (16,99.46)};
    \addplot[black,thick,dash pattern=on 10pt off 5pt]
      coordinates {(1,9.80) (16,9.80)};
    \legend{QLoRA,Baseline}
  \end{axis}
\end{tikzpicture}
  \end{minipage}
  \begin{minipage}{0.48\linewidth}
    \begin{tikzpicture}
      \begin{axis}[
        title={\textbf{b. GSM8k (source)}},
        xlabel={Rank}, xtick={1,4,8,16},
        ylabel={Accuracy (\%)},
        ymin=10, ymax=25,
        legend pos=north east,
        legend style={font=\footnotesize},
        width=\textwidth,
      ]
        \addplot[red,very thick,mark=*] coordinates {(1,11.07) (4,13.50) (8,13.57) (16,14.18)};
        \addplot[black,thick,dash pattern=on 10pt off 5pt]             coordinates {(1,19.94) (16,19.94)};
        \legend{QLoRA,Baseline}
      \end{axis}
    \end{tikzpicture}
  \end{minipage}\vspace{1em}
  \begin{minipage}{0.48\linewidth}
    \begin{tikzpicture}
      \begin{axis}[
        title={\textbf{c. HellaSwag (source)}},
        xlabel={Rank}, xtick={1,4,8,16},
        ylabel={Accuracy (\%)},
        ymin=49, ymax=53,
        legend pos=north east,
        legend style={font=\footnotesize},
        width=\textwidth,
        yticklabel style={
          /pgf/number format/fixed,       
          minimum width=3em,              
          align=right                     
        },
      ]
        \addplot[orange,very thick,mark=*] coordinates {(1,49.71) (4,49.64) (8,49.86) (16,50.09)};
        \addplot[black,thick,dash pattern=on 10pt off 5pt]                coordinates {(1,50.85) (16,50.85)};
        \legend{QLoRA,Baseline}
      \end{axis}
    \end{tikzpicture}
  \end{minipage}\hfill
  \begin{minipage}{0.48\linewidth}
    \begin{tikzpicture}
      \begin{axis}[
        title={\textbf{d. PIQA (source)}},
        xlabel={Rank}, xtick={1,4,8,16},
        ylabel={Accuracy (\%)},
        ymin=73, ymax=77,
        legend pos=north east,
        legend style={font=\footnotesize},
        width=\textwidth,
      ]
        \addplot[green!60!black,very thick,mark=*] coordinates {(1,73.94) (4,74.10) (8,74.54) (16,74.27)};
        \addplot[black,thick,dash pattern=on 10pt off 5pt]               coordinates {(1,75.19) (16,75.19)};
        \legend{QLoRA,Baseline}
      \end{axis}
    \end{tikzpicture}
  \end{minipage}
  \caption{
  QLoRA experiments, with \texttt{unsloth/SmolLM2-1.7B-Instruct-bnb-4bit} as the base model.
  We report the test accuracy as a function of the QLoRA rank on the source (\textbf{a.}) and target tasks (\textbf{b.-d.}).
  Note that the baseline performance is lower than on \cref{fig:smollm2_forgetting_step} due to the impact of 4-bit quantization of the base model.
  \textbf{a.}: In spite of performance of the original model being around chance level (\cref{tab:smollm2_mnist_baseline}), our Textual MNIST task is learnable across all tested QLoRA ranks (1-16).
  \textbf{b.-d.}: Performance on \emph{source} tasks appears to increase with the QLoRA rank.
  A higher rank map to higher trainable parameter counts; thus, it could be expected to lead to a more significant deviation from the base model, more pronounced overfitting to the Textual MNIST task, and thus \emph{worse} performance degradation on the source tasks.
  However, this is not what we observe.
  A low-rank $\Delta W$ might be counter-productive: while it enforces a lower the number of trainable parameters, it does so in a rigid way.
  In contrast, a higher-rank $\Delta W$ might more readily express a more "natural",  less ad-hoc adaptation of the base model's weights.
  This effect, combined with the regularization induced by weight decay, might lead to the enhanced preservation of source-task performance with increasing rank that we observe.
  }
  \label{fig:lora_rank}
\end{figure*}

\subsection{Learning Textual MNIST}
\label{learning_textual_mnist}

In Figure~\ref{fig:smollm2_forgetting_step}.a, we report the test accuracy on Textual MNIST evaluated at each iteration before pruning, and compare it to that of the random mask.
After the first step, which is similar to full fine-tuning since the adapter hasn't been pruned yet, the model achieves a 99.05 \% test accuracy on Textual MNIST. Despite having over 1.6B free parameters for a dataset of only 60,000 examples, the model still manages to generalize to the test set. At each subsequent step, we observe some performance degradation, as expected, almost dropping to the performance of the random mask at step 4. We hypothesize this might be due to the high pruning ratio per iteration (80 \%), or poor tuning of other hyperparameters.

\subsection{Quantifying Forgetting}
\label{forgetting}

So far, we have found a sparse subnetwork $\Delta W_T$ and used it to adapt SmolLM2 to the Textual MNIST target task.
How did this adaptation affect the ability of the model to perform other tasks?
First, we qualitatively verify that we can still chat with the model, and subsequently ask it to classify a digit (Figure~\ref{conversation}).
Then, we quantitatively evaluate forgetting using the \emph{source tasks} defined in \cref{source_tasks}, by comparing its initial accuracy on those tasks to its accuracy after being fine-tuned with our method on the \emph{target} task.

We reproduce the evaluation of SmolLM2-1.7B-Instruct on those benchmarks using the LM Evaluation Harness~\citep{eval-harness}, including the 4-bit quantized version of the model, and present our results in Table~\ref{tab:smollm2_forgetting_baseline}.
Interestingly, quantization seems to have a significant effect on GSM8k, with accuracy dropping from 43.7 \% to 19.9 \%, but not on PIQA and HellaSwag.

We evaluate our sparse adapter at each pruning iteration on GSM8k (5-shot), HellaSwag (0-shot), and PIQA (0-shot), and present the results respectively in Figures~\ref{fig:smollm2_forgetting_step}.b,~\ref{fig:smollm2_forgetting_step}.c, and~\ref{fig:smollm2_forgetting_step}.d, along with the full-precision baseline of \cref{tab:smollm2_forgetting_baseline}.
For HellaSwag and PIQA, we don't observe significant forgetting at any pruning iteration. However, for GSM8k, the dense adapter's accuracy (first pruning iteration) drops to 40.7~\% from our baseline of 43.7~\%. As it becomes sparse, accuracy is regained and ultimately outperforms our baseline, reaching 45.1~\% at step 3. While this result could suggest that sparse adaptation on Textual MNIST boosts accuracy on mathematical tasks, the model still doesn't reach the accuracy of 48.8~\% reported by the authors of SmolLM2~\citep{allal2025smollm2}. This could be better explained by the inter-run variability of the benchmarks. Without multiple runs, we cannot definitively attribute the improvements to our method rather than measurement noise, but running each benchmark multiple times was not feasible within our compute budget.

\subsection{Comparison with QLoRA}

We demonstrated the feasibility of finding a sparse adapter to adapt SmolLM2 to a new target task, and investigated the effect of that adaptation on forgetting.
But how does our method compare to other parameter-efficient fine-tuning techniques?

QLoRA~\citep{dettmers2024qlora} is a popular, memory efficient technique for fine-tuning LLMs that applies a LoRA~\citep{hu2021lora} adapter on top of a 4-bit quantized model. We fine-tune a 4-bit quantized version of SmolLM2-1.7B-Instruct from Unsloth~\citep{unsloth} on Textual MNIST using QLoRA. We perform this experiment with four different ranks, as the rank is proportional to the number of free parameters. We use the AdamW~\citep{loshchilov2018decoupled} optimizer, and report our hyperparameters in Table~\ref{tab:qlora_hp}. 

\begin{table}
  \caption{Hyperparameters for QLoRA adaptation of SmolLM2-1.7B-Instruct.}
  \label{tab:qlora_hp}
  \centering
  \begin{tabular}{lc}
    \toprule
    \textbf{Hyperparameter} & \textbf{Value} \\
    \midrule
    Rank          & 1/4/8/16 \\
  Alpha    & 32 \\
    Dropout             & 0.0 \\
    Learning Rate       & 2e-4 \\
    Scheduler           & Linear \\
    Weight Decay        & 0.1 \\
    Batch Size          & 16 \\
    Warmup Steps        & 1000 \\
    Epochs              & 4 \\
    \bottomrule
  \end{tabular}
\end{table}

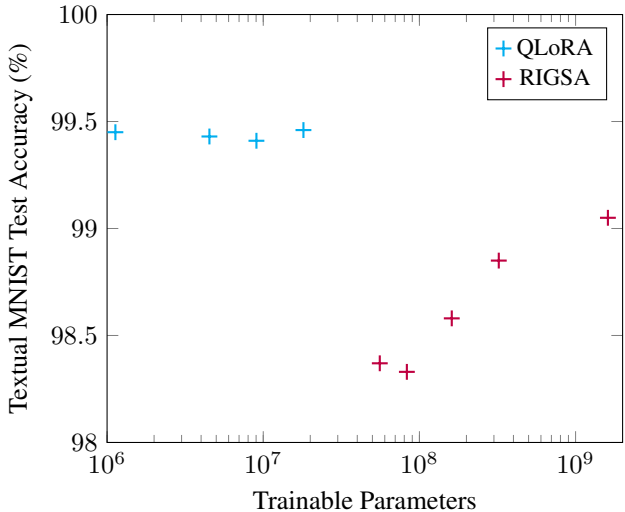
\begin{figure}
    \centering
\begin{tikzpicture}
  \begin{axis}[
    xlabel={Trainable Parameters},
    ylabel={Textual MNIST Test Accuracy (\%)},
    ymin=98, ymax=100,
    xmin=1e6, xmax=2e9,
    xmode=log,
    log basis x=10,
    legend pos=north east,
    legend style={font=\footnotesize},
    mark size=3pt
  ]
    \addplot[
      only marks,
      mark=+,
      mark options={thick},
      cyan
    ] coordinates {
      (1130496, 99.45)
      (4521984, 99.43)
      (9043968, 99.41)
      (18087936, 99.46)
    };

    \addplot[
      only marks,
      mark=+,
      mark options={thick},
      purple
    ] coordinates {
      (55746407, 98.37)
      (83103630, 98.33)
      (161125260, 98.58)
      (322269459, 98.85)
      (1610612736, 99.05)
    };

    \legend{QLoRA, RIGSA}
  \end{axis}
\end{tikzpicture}
  \caption{
  Test accuracy on Textual MNIST vs. number of trainable parameters across adapters.
  Test accuracy for QLoRA remains relatively stable across trainable parameter budgets.
  In contrast, RIGSA's performance increases with the parameter budget.
  }
  \label{fig:accuracy_parameters}
\end{figure}

We then evaluate the test accuracy of each model on Textual MNIST, and report it in Figure~\ref{fig:lora_rank}.a. Additionally, we also run the source task benchmarks for GSM8k, HellaSwag, and PIQA, and compare the performance to the quantized baseline we obtained in \cref{tab:smollm2_forgetting_baseline}, and report our findings respectively in Figures~\ref{fig:lora_rank}.b,~\ref{fig:lora_rank}.c, and~\ref{fig:lora_rank}.d. We find that QLoRA outperforms sparse fine-tuning on the target task, achieving 99.46 \% test accuracy on Textual MNIST when the rank is 16, where our sparse adapter achieved only 98.37~\%. Interestingly, it also outperforms the 99.05~\% achieved with dense fine-tuning in this setting.
This might indicate superior regularization capabilities of QLoRA, which allowed us to train for 4 epochs instead of 1 without observing over-fitting.
To better compare sparse adaptation and QLoRA, we plot in Figure~\ref{fig:accuracy_parameters} the test accuracy against the number of trainable parameters, and observe that sparse adaptation consistently underperforms QLoRA on Textual MNIST.

In terms of forgetting, however, QLoRA seems to forget significantly more than our sparse adaptation method, resulting in a drop from 19.9 \% accuracy of the quantized baseline on GSM8k to 14.18 \% when the rank is 16, and down to 11.07 \% when the rank is 1.

\section{Discussion}
\label{discussion}

We proposed the RIGSA method, as well as the Textual MNIST task, which is meant to be easy for small language models to learn through fine-tuning yet challenging without fine-tuning.
We applied our RIGSA method to the fine-tuning of SmolLM2-1.7B-Instruct on Textual MNIST, and evaluating the model's accuracy on standard benchmarks before and after fine-tuning.

Our results clearly show that SmolLM2-1.7B-Instruct initially struggles to classify Textual MNIST (\cref{tab:smollm2_mnist_baseline}), but learns this task after just one epoch of training with random masking or RIGSA (\cref{fig:smollm2_forgetting_step}).
The model also learns Textual MNIST well via QLoRA (\cref{fig:lora_rank}).

While RIGSA achieves lower final accuracy than QLoRA on the target task (98.37\% vs 99.46\%), it demonstrates superior resistance to catastrophic forgetting, particularly on GSM8k where QLoRA's performance drops to 14.18\% from the 19.9\% baseline.
RIGSA does not conclusively outperform random masking, with both methods leading to modest and comparable forgetting on source tasks (\cref{fig:smollm2_forgetting_step}).
The unexpected improvement on source tasks with random masking warrants further investigation with proper statistical analysis.
Due to compute budget limitations and the inherent variability of benchmarks, we cannot definitively attribute these improvements to our method rather than measurement noise.

An interesting observation does emerge from even such limited experiments.
RIGSA mobilizes significantly more trainable parameters (\cref{fig:accuracy_parameters}) than QLoRA across the range of iterations that we tested; one might thus expect it to overfit to the target task, and display worse performance degradation on source tasks.
However, \textbf{this is not the case}.
While QLoRA achieves higher accuracy on the target task, RIGSA displays considerably less forgetting than QLoRA in spite of having more trainable parameters.
This effect is clear on all three benchmarks under consideration (PIQA, HellaSwag, and GSM8k), and applies to random masking as well, suggesting that sparse adaptation may offer inherent regularization benefits.

We acknowledge that any definitive conclusions as to the effectiveness of RIGSA compared to random masking will require more extensive hyperparameter search, as well as multiple runs of the GSM8k, HellaSwag and PIQA benchmarks.
In particular, we encourage future work to consider the impact of the masking ratio (including higher masking ratios, in order to match the trainable parameter counts of common (Q)LoRA configurations), a more direct comparison with LT-SFT, and a comparison with single-precision LoRA rather than only 4-bit QLoRA.

\section*{Impact Statement}

The weights of foundation models are obtained through the long and expensive process of pre-training.
This represents a tremendous investment of human time, electrical power, and dedicated computing hardware.
As foundation models and derivative works multiply, making the best use of pretrained weights is of increasing importance.
In line with research on parameter-efficient fine-tuning at large, our work aims to support the development of downstream applications from foundation model weights using less data and less computational power than naive alternatives.

\bibliography{bibliography}
\bibliographystyle{icml2025}

\newpage
\appendix
\onecolumn

\begin{figure*}
\centering
\ttfamily\tiny
\begin{minipage}{0.8\textwidth}
\begin{verbatim}
<|im_start|>system
You are a helpful AI assistant named SmolLM, trained by Hugging Face<|im_end|>
<|im_start|>user
Below is a digit as text. Which digit is it?
0000000000000000000000000000
0000000000000000000000000000
0000000000000000000000000000
0000000000000000000000000000
0000000000000002592000000000
0000000000000068998200000000
0000000000001399999920000000
0000000000007999959960000000
0000000000029964402893000000
0000000000289920000697000000
0000000000999700000499000000
0000000000997100000499000000
0000000017994000000499000000
0000000039990000000099000000
0000000099930000000099000000
0000000499710000000099000000
0000001799400000000299000000
0000003999000000006895000000
0000007999000000379962000000
0000003999000004999500000000
0000003999335999970000000000
0000001799999999710000000000
0000000099999986000000000000
0000000033733320000000000000
0000000000000000000000000000
0000000000000000000000000000
0000000000000000000000000000
0000000000000000000000000000
<|im_end|>
<|im_start|>assistant
The digit is 
\end{verbatim}
\end{minipage}
\caption{Textual MNIST classification prompt.}
\label{fig:textual_mnist_prompt}
\end{figure*}

\begin{figure*}
  \centering
  \fbox{\includegraphics[width=0.6\textwidth]{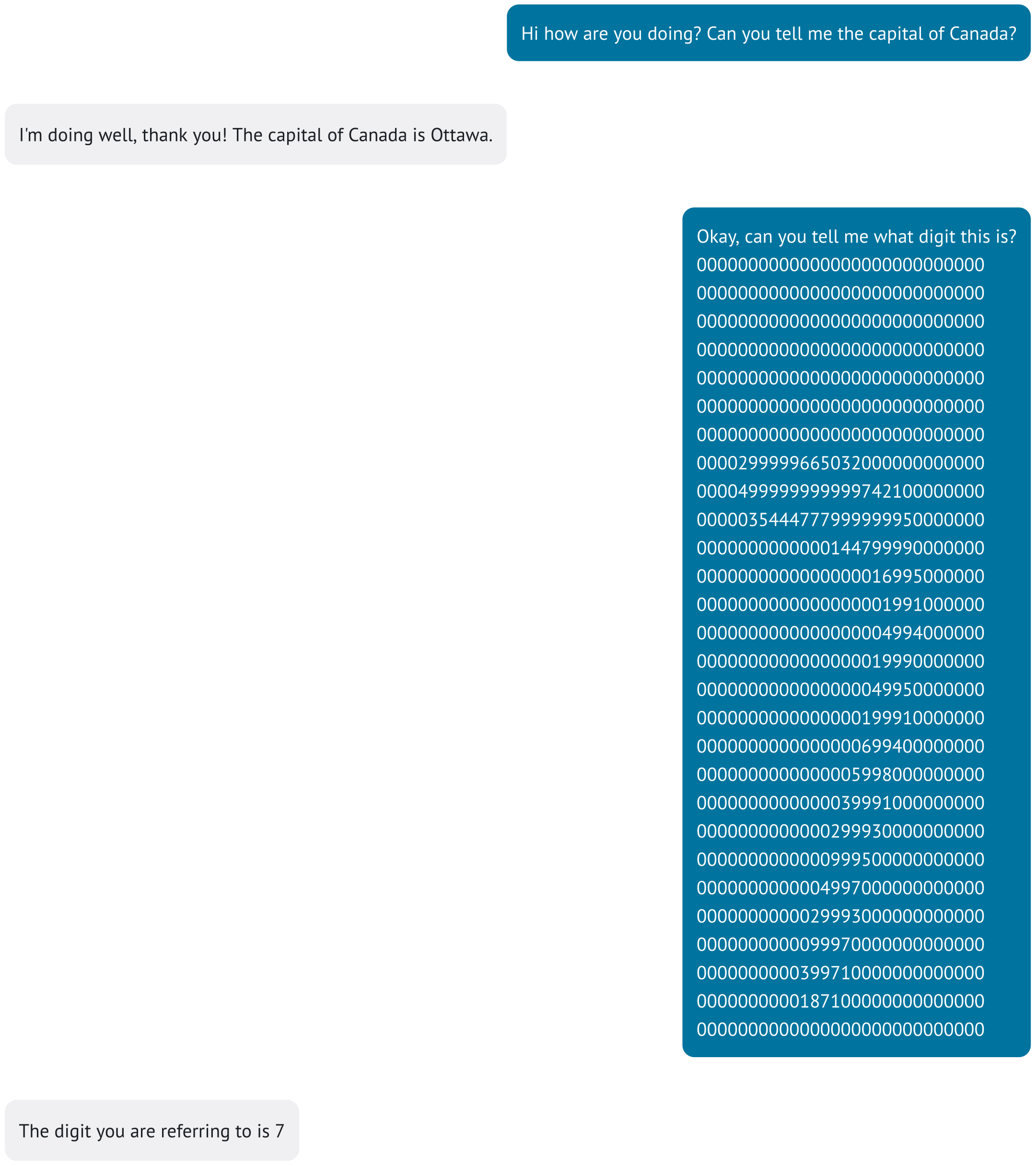}}
  \caption{Conversation with SmolLM2-1.7-Instruct adapted to Textual MNIST.}
    \label{conversation}
\end{figure*}

\end{document}